\documentclass[final,12pt]{clear2025} %

\usepackage[utf8]{inputenc} %
\usepackage[T1]{fontenc}    %
\usepackage{hyperref}       %
\usepackage{url}            %
\usepackage{booktabs}       %
\usepackage{amsfonts}       %
\usepackage{nicefrac}       %
\usepackage{microtype}      %
\usepackage{float}
\usepackage{subcaption}
\usepackage{array}
\usepackage{comment}
\usepackage[font=small,labelfont=bf,tableposition=top]{caption}

\usepackage[subtle, mathdisplays=normal]{savetrees}

\usepackage{graphicx}
\usepackage{amssymb}
\usepackage{multirow}

\usepackage[toc,page,header]{appendix}
\usepackage{minitoc}

\renewcommand \thepart{}
\renewcommand \partname{}

\hypersetup{ 
	pdftitle={Causal Repr Learning},
	pdfkeywords={},
	pdfborder=0 0 0,
	pdfpagemode=UseNone,
	colorlinks=true,
	linkcolor=blue, %
	citecolor=blue, %
	filecolor=blue, %
	urlcolor=blue, %
	pdfview=FitH,
	pdfauthor={Anonymous},
}

\newcommand\GeneA{Gene \textit{A}}
\newcommand\GeneB{Gene \textit{B}}

\newcommand{\github}{\href{https://github.com/kurowasan/cd_datasets}{github repo}}

\renewcommand \thepart{}
\renewcommand \partname{}

\title[Grounding Causal Discovery in Real-World Applications]{The Landscape of Causal Discovery Data: \\
Grounding Causal Discovery in Real-World Applications}
\usepackage{times}

\clearauthor{%
 \Name{Philippe Brouillard} \Email{philippebrouillard@gmail.com}\\
 \addr Mila-Québec, Université de Montréal
 \AND
 \Name{Chandler Squires} \Email{csquires@andrew.cmu.edu}\\
 \addr Carnegie Mellon University
 \AND
 \Name{Jonas Wahl} \Email{jonas.wahl@dfki.de}\\
 \addr Deutsches Forschungszentrum für k\"unstliche Intelligenz (DFKI)
 \AND
 \Name{Konrad P. K\"ording} \Email{kording@upenn.edu}\\
 \addr University of Pennsylvania
 \AND
 \Name{Karen Sachs} \Email{karen@nextgenanalytics.us}\\
 \addr Next Generation Analytics and Modulo Bio
 \AND
 \Name{Alexandre Drouin$^\dagger$} \Email{alexandre.drouin@servicenow.com}\\
 \addr ServiceNow Research, Mila-Québec, Université Laval
 \AND
 \Name{Dhanya Sridhar$^\dagger$} \Email{dhanya.sridhar@mila.quebec}\\
 \addr Mila-Québec, Université de Montréal\\[3mm]
 $^\dagger$ Equal supervision
}

\begin{document}

\maketitle

\begin{abstract}
  Causal discovery aims to automatically uncover causal relationships from data, a capability with significant potential across many scientific disciplines. However, its real-world applications remain limited. Current methods often rely on unrealistic assumptions and are evaluated only on simple synthetic toy datasets, often with inadequate evaluation metrics. In this paper, we substantiate these claims by performing a systematic review of the recent causal discovery literature. We present applications in biology, neuroscience, and Earth sciences—fields where causal discovery holds promise for addressing key challenges. We highlight available simulated and real-world datasets from these domains and discuss common assumption violations that have spurred the development of new methods. Our goal is to encourage the community to adopt better evaluation practices by utilizing realistic datasets and more adequate metrics. 

\end{abstract}


\section{Introduction} \label{sec:intro}
    In many scientific endeavors, researchers are not merely interested in identifying statistical patterns, but in understanding the underlying causal relationships that govern complex systems. They want to answer causal questions such as ``What would be the impact of changing a specific variable on this system?''. This kind of question cannot be answered by purely statistical models. For instance, in healthcare, understanding causal relationships is essential to determine the efficacy of treatments leading to better patient outcomes and more efficient resource allocation. If a purely statistical model is used instead, the model might rely on spurious correlations, leading to erroneous conclusions. Causal discovery aims at recovering causal relations directly from data, allowing us to answer causal queries. While causal inference is challenging, most scientific fields could benefit from that capability. 
    
    That being said, the field of causal discovery is predominantly method-driven rather than application-driven \citep{rolnick2024application}: the community produces new methods and algorithms at high speed but still relies on toy datasets and simple metrics for their evaluation \citep{gentzel2019case}, impeding its development and applicability to real-world problems. Recently, a plethora of surveys of causal discovery have covered existing causal discovery methods \citep{wang2024survey, hasan2023survey, zanga2022survey, assaad2022survey, vowels2022d, zhou2022survey, nogueira2021causal, guo2020survey, glymour2019review, malinsky2018causal, singh2017comparative}, but none focused on the datasets and real-world applications to which these methods were applied. 
    However, using good datasets and benchmarks is just as crucial as having good algorithms. For example, this has been pivotal in the recent deep learning boom with datasets such as ImageNet \citep{deng2009imagenet} and its associated challenge \citep{russakovsky2015imagenet}. Beyond the choice of datasets, there is also a need for deeper consideration of the types of problems to which causal discovery can and should be applied. Over-reliance on simple settings makes the field disconnected from real-world challenges, and without practical applications, causal discovery risks becoming merely theoretical storytelling.

    The goal of this review is to incite the community to be more application-driven: we do that by surveying the recent literature and highlighting key methodological shortcomings to be improved, as well as identifying fields that seem ripe to benefit the application of causal discovery. First, by performing a systematic review, we show in Section~\ref{sec:survey} that the field of causal discovery still relies on synthetic datasets and a low diversity of real-world datasets. Also, in most studies, inadequate metrics are used for evaluation. Second, in Section~\ref{sec:unique_challenges}, we show that many alternatives exist to simple synthetic datasets, both pseudo-real and real-world datasets, and we provide a list of some common datasets that are used to assess new causal discovery methods (see our \github). Finally, we highlight a few key scientific fields --- biology, neuroscience, and Earth sciences --- where a significant amount of real-world data is generated and causal discovery should be a short-term target. 
    Overall, the resulting overview reveals that real-world applications frequently challenge established causal discovery assumptions and may serve as catalysts for innovation, underscoring the importance of grounding research in practical scenarios and utilizing real-world datasets over purely synthetic ones.

\section{Background} \label{sec:background}
    Causal models make formal predictions about the effects of intervention, i.e., external manipulations that set a variable to some specific value or distribution. While many approaches exist to this end, this section briefly presents one specific class of causal models that is popular in the field. We detail the entailed assumptions and introduce causal discovery (for details, see \citet{peters2017elements}; \citet{pearl2009causality}).

    \textbf{Causal Bayesian Network.} A \textit{Causal Bayesian Network} (CBN) consists of a \textit{directed acyclic graph} (DAG) $G = (V, E)$ with $|V| = d$ and a random vector $X = (X_1, \dots, X_d) \sim P_X$ whose entries correspond to the nodes of $G$. The distribution $P_X$ is connected to the graph $G$ by the Markov property which asserts that $P_X$ factorizes as
    \begin{equation} \label{eq:joint_obs}
        P_X = \prod_{i = 1}^d P^{(0)}(X_i \mid pa_i^G),
    \end{equation}
    where $pa_i^G$ are the parents of $X_i$ in the graph $G$. Up to this point, this model is a standard Bayesian network. The causal semantics stem from the interventional interpretation of the edge directions and the fact that interventions on variables can also be considered. Let $(I_1, \dots, I_k)$ be a collection of interventional targets. Each interventional target $I_j \subseteq [d]$ represents a set of variables that have been intervened upon during intervention $j$. The distribution induced by the $j$-th intervention is given by
    \begin{equation}
        P^{(j)}_X = \prod_{i \notin I_j}P^{(0)}(X_i \mid pa_i^G)\prod_{i \in I_j}P^{(j)}(X_i \mid pa_i^G),
    \end{equation}
    where $P^{(0)}$ are the observational conditionals that stay invariants (i.e., same as in Eq~\ref{eq:joint_obs}) and $P^{(j)}$ are conditionals intervened upon which are specific to the interventional distribution $j$. This is a general formulation, with perfect interventions being a notable specific case that corresponds to a setting where the conditional $P^{(j)}(X_i)$ does not depend on its parents $pa^G_i$.
    
    \textbf{Causal discovery.} The task of recovering the graph $G$ from a dataset $\mathcal{D}$ (possibly containing interventional data) is called causal discovery. Constraint-based methods such as the PC algorithm \citep{spirtes2001causation} perform conditional independence tests to recover $G$, while score-based methods achieve this by finding the graph that maximizes a score, such as the Bayesian Information Criterion (BIC) \citep{chickering2002optimal}. Some hybrid methods combine aspects of both approaches. Other methods make parametric assumptions on the functional form of the causal mechanisms or the variable distributions and orient edges based on detected asymmetries. For a more complete presentation, see for example \citet{glymour2019review}. All of these methods are only guaranteed to recover the correct ``ground-truth'' graph in the infinite sample limit if the data-generating mechanism satisfies specific assumptions.

    \textbf{Common assumptions.} Causal discovery relies on many assumptions, some directly induced by the CBN approach: 1) acyclicity of the graph over variables, 2) \textit{causal sufficiency} which refers to the fact that there are no unobserved confounders, i.e., variables that are parents to more than one $X_i$, 3) the \textit{faithfulness} assumption that stipulates that conditional independencies in the distribution $P_X$ implies the corresponding d-separation in the graph $G$, and 4) the random variables provide an appropriate representation to reason about the problem of interest \citep{spirtes2009variable, eberhardt2016green}.
    As already mentioned above, many methods also assume a particular functional form of the causal mechanisms (e.g., linearity). In practice, as we will explain in more detail in Section~\ref{sec:challenges}, most of these assumptions are violated in real-world problems. 
    
    Even when all these assumptions are satisfied, causal discovery is a hard task, both combinatorically and statistically. The space of DAGs scales super-exponentially with respect to the number of variables, and the assumptions above only guarantee correctness in the infinite sample limit, while in practice, one also has to deal with finite sample errors. Moreover, when the data is purely observational, without further assumptions one can at best identify an equivalence class of graphs, called the Markov Equivalence Class. Utilizing interventions represents the optimal strategy for overcoming obstacles in causal identifiability since it can greatly shrink the size of the equivalence class. If single-target interventions are performed on every node except one, the ground-truth graph is identifiable and, in general, fewer interventions are required  \citep{eberhardt2012number}.

    \textbf{Evaluation.} To evaluate the performance of causal discovery algorithms, there are, broadly speaking, four classes of metrics: structural, qualitative, observational, and interventional. \textit{Structural} metrics consist of comparing the learned graph to the ground-truth graph using distances, such as the \textit{structural Hamming distance} (SHD) which counts the total number of edges that are missing, superfluous, and reversed. \textit{Qualitative} assessments consist of experts in the field who will discuss, based on their domain knowledge, the plausibility of some causal relations. This is similar to structural measure, but it is used when the ground-truth graph is not known. \textit{Observational} and \textit{interventional} metrics correspond to evaluating how well the learned model predicts held-out observational data and data from an unseen intervention, respectively. The latter is arguably closest to what most practitioners care about: the ability to predict the effect of unseen interventions. Appendix~\ref{app:metrics} provides further details on these metrics.

\section{From Purely Synthetic Datasets to Real-World Datasets} \label{sec:datasets}
We first describe different families of datasets that are available for evaluating causal discovery methods before analyzing their use in recent papers.

\subsection{Synthetic Datasets} \label{subsec:synth_data}
    We need to address the elephant in the room: many causal discovery methods are evaluated only on simple synthetic datasets that do not reflect any real-world phenomenon \citep{gentzel2019case} (a claim that we will also demonstrate in Section~\ref{sec:survey}). Moreover, many of these synthetic datasets even use exactly the same generator as the model fitted. Still, synthetic datasets are used since they offer many advantages: the ground-truth causal graph is known, a large sample size can be used and different properties of the causal model can be precisely controlled (e.g. density of the graph, number of vertices, functional form, etc) to assess a method. By design, the generated data will perfectly respect many stringent assumptions such as causal sufficiency, faithfulness, a particular functional form, etc. Moreover, a motivation for using synthetic datasets might come from the misconception that real-world datasets are scarce or impossible to evaluate quantitatively, a notion we aim to refute.

    Synthetic datasets are usually generated by following these steps: first, the causal graph is sampled, then the causal mechanisms parameters, and finally, the data is sampled using ancestral sampling (i.e., by sampling the variables following their topological ordering). Common approaches to sample a DAG include the Erd\H{o}s-R\'enyi scheme, which samples directed edges with a fixed probability, and scale-free networks \citep{barabasi1999emergence}, which samples edges following a preferential attachment process  \citep{barabasi2009scale}. The causal mechanisms parameters often follow a particular functional form assumed by the causal discovery method. For instance, one of the most common causal mechanisms is linear relations with Gaussian noise. Alternatives include nonlinear~\citep{peters2014causal} and post-nonlinear additive noise models~\citep{zhang2012identifiability} which are often used since they lead to identifiability results.

    However, as highlighted in \citet{reisach2021beware} and \citet{reisach2024scale}, the way these datasets are generated can sometimes be problematic as it introduces artifacts that some causal discovery methods may exploit. Namely, \citet{reisach2021beware} showed that one can recover the causal ordering of some synthetic datasets simply by sorting the variables according to their variances. Even when these pitfalls are avoided by changing the data generating process \citep{andrews2024better,ormaniec2024standardizing}, synthetic datasets are much simpler than their real-world counterpart and thus, the performance of proposed causal discovery methods is overestimated \citep{eigenmann2020evaluation}. As we will elaborate in Section~\ref{sec:challenges}, real-world problems rarely conform to many of the assumptions built into synthetic datasets. We finish by noting that synthetic datasets can still be used for more realistic evaluation by benchmarking causal discovery methods in scenarios where these stringent assumptions are violated, as in \citet{montagna2024assumption}.

\subsection{Real-World Datasets} \label{subsec:realworld}
    In the end, what we really care about is the application of causal discovery to real-world problems. Real-world datasets are particularly interesting since they often break common assumptions and inform the causal discovery community of what are remaining and interesting challenges to overcome. The primary limitation of real-world datasets, compared to synthetic and pseudo-real datasets, is the difficulty in evaluating the quality of discovered structures due to the absence of a known ground truth, making most assessments qualitative. However, when interventional data are available, we stress the fact that quantitative evaluation is possible via interventional metrics. 

    \begin{figure}[t] 
        \includegraphics[width=0.8\textwidth]{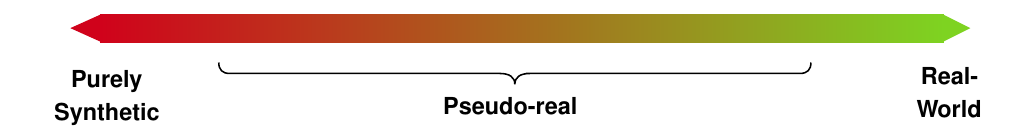}
        \centering
        \caption{The realism of datasets spans a spectrum, from purely synthetic to real-world datasets.}
        \label{fig:datasets_realism}
        \vspace{-6pt}
    \end{figure}
    
\subsection{Pseudo-Real Datasets} \label{subsec:pseudoreal}
    Pseudo-real datasets are designed to be similar to real-world data while retaining the benefits of synthetic datasets: a known ground-truth graph, adherence to common assumptions, and control over generation parameters. As a result, some strongly advocate to use this type of data \citep{glymour2019evaluation}.

    Many pseudo-real datasets rely on a data generation process inspired by mathematical models, such as ordinary or stochastic differential equations, used in their respective fields. In biology, many simulators that generate synthetic gene expression data have been proposed: \emph{SynTReN} \citep{van2006syntren}, \emph{GeneNetWeaver} \citep{schaffter2011genenetweaver}, \emph{BEELINE} \citep{pratapa2020benchmarking}, \emph{SERGIO} \citep{dibaeinia2020sergio}. Similarly, in neuroscience, various simulators have been proposed such as simulated spiking interactions between neurons from hippocampus \citep{bezaire2015modeling}, simulated network dynamics between network areas approximated by mean-field dynamics along with fMRI signal generation and calibrated against some brain data \citep{smith2011network}. Additionally, several datasets derive from variants of the virtual brain project \citep{sanz2013virtual}. In  Earth sciences, \citet{ebert2017causal} simulate data that reflects typical advection and diffusion processes in the planet's atmosphere to investigate unexplained connections found by causal discovery algorithms on real-world data.

    Alternatively, some pseudo-real datasets are generated by directly learning a model from real-world datasets. Once fitted, the model can produce examples similar to the original data under different conditions, with the model's graph serving as the ground truth. This often leads to datasets that, by design, respect most common causal discovery assumptions. The most popular resource of that type is the \href{https://www.bnlearn.com/bnrepository/}{bnlearn repository} \citep{scutari2009learning, friedman1997impact} that contains several datasets from a wide range of fields. More recently, in the medical setting, \citet{tu2019neuropathic} created a simulator for neuropathic pain diagnosis.
    In the field of manufacturing \citep{vukovic2022causal}, \citet{gobler2023causalassembly} proposed a Benchmark called \emph{Causalassembly} where a model has been fitted to real production line data.
    \citet{runge2019inferring} propose the platform \textit{CauseMe} that contains several time-series datasets, some pseudo-real and some real with a consensus graph. Finally, \citet{lawrence2021data, cheng2023causaltime} propose more general frameworks where several datasets from different fields can be combined for generating realistic time-series data. Simply put, their methods can be applied to any real-world dataset and yield a new simulator.
    
    The realism of datasets can be viewed as a spectrum, with purely synthetic datasets at one end and real-world datasets at the other (see Fig.~\ref{fig:datasets_realism}). Pseudo-real datasets fall in between and they can greatly vary in their realism: on one end, they can resemble purely synthetic datasets by respecting all common assumptions and integrating little information from real-world data (e.g., only the graph), at the other end, they can represent a significant improvement over synthetic datasets as they resemble real-world problems and can violate common causal assumptions. As an example of the latter, the simulator of \citet{smith2011network} generates cycles and the simulator of \citet{tu2019neuropathic} can generate data with unknown confounders, selection bias, and missing data. In short, a good simulator should faithfully replicate real-world datasets in all their complexity and, hopefully, causal discovery methods that perform well on the simulator should also transfer to real-world datasets. To do so, considering real-world problems and datasets is essential when designing pseudo-real datasets to ensure their realism and practical relevance.

\section{Systematic Review of the Causal Discovery Literature} \label{sec:survey}
    To better understand the trends in the causal discovery community regarding dataset use and evaluation metrics, we conducted a systematic literature review similar to the one of \citet{gentzel2019case}. Using the Semantic Scholar API~\citep{semanticscholar}, we collected scientific articles on causal discovery published between 2019 and 2024 at major machine learning conferences (NeurIPS, ICLR, ICML, AISTATS, UAI, AAAI, and CLeaR). We collected a total of 221 papers and, after manually filtering them, we retained 167 papers. A detailed presentation of our methodology, along with some additional results, can be found in Appendix~\ref{app:systematic_review}. The list of selected papers and the analysis code are available at our \github. 

\begin{figure}
      \begin{minipage}[b]{.3\linewidth}
        \centering
        \includegraphics[width=0.8\linewidth]{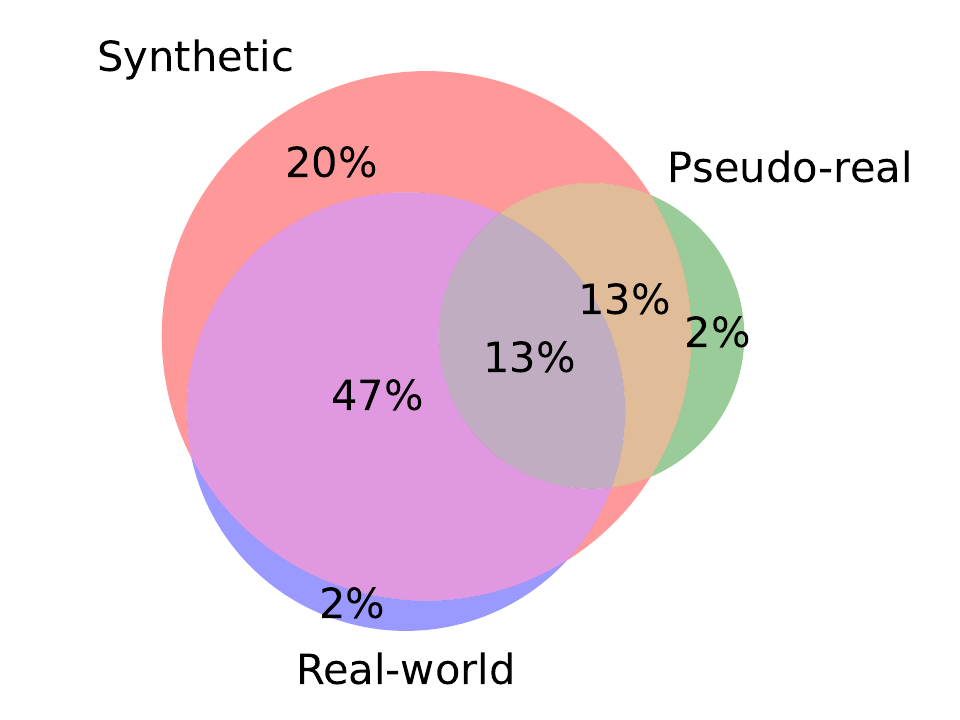} 
        \captionof{figure}{Distribution of papers based on used dataset types.}
        \label{fig:venn}
      \end{minipage}\hfill
      \begin{minipage}[b]{.33\linewidth}
        \centering
        \includegraphics[width=0.9\linewidth]{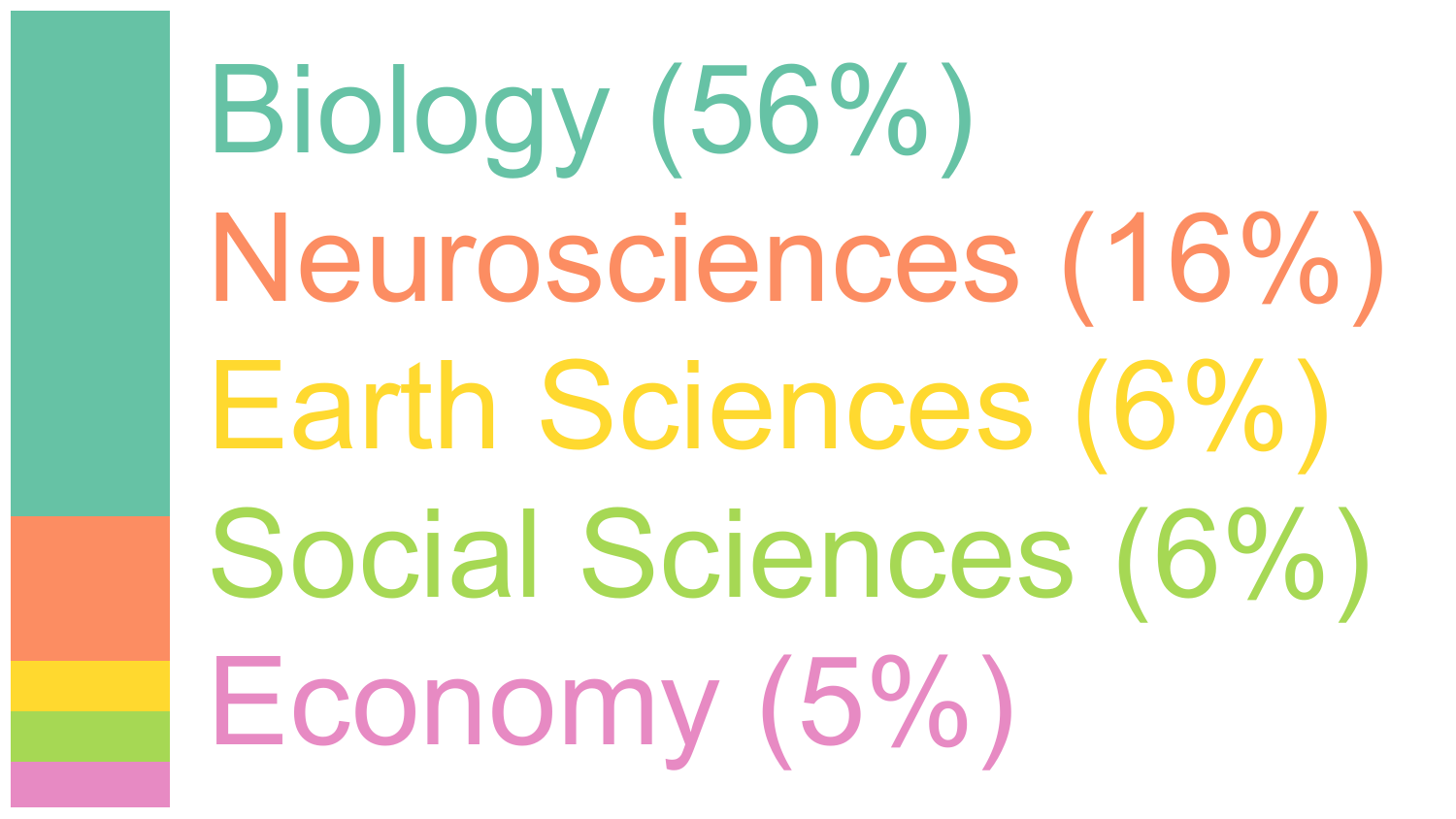}
        \captionof{figure}{Common fields of the pseudo-real and real-world datasets.}
        \label{fig:fields}
      \end{minipage}\hfill
      \begin{minipage}[b]{.3\linewidth}
        \centering
        \scriptsize
        \begin{tabular}{l|r|r|r}
            & &Real- \\
            &Simulated &world \\\midrule
            Structural &100.0\% &67.3\% \\
            Qualitative &3.0\% &36.4\% \\
            Observational &5.5\% &5.6\% \\
            Interventional &9.1\% &7.5\% \\
            \bottomrule
        \end{tabular}
        \captionof{table}{Percentage of studies using evaluation metrics.} 
        \label{tab:metrics}
      \end{minipage}\hfill
    \end{figure}

    Fig~\ref{fig:venn} shows the distribution of dataset types used in the selected studies.
    We observe that 20\% of these only make use of purely synthetic datasets, while 62\% rely on real-world datasets.
    Most real-world datasets are small, with 80\% containing 20 or fewer variables. Fig~\ref{fig:fields} shows the field of provenance of pseudo-real and real-world datasets. Biology is by far the most prevalent field. This is partly explained by the ubiquitous use of the flow cytometry dataset, often simply named \textit{Sachs} \citep{sachs2005causal}.  It is the only real-world dataset considered for 35\% of all the papers relying on real-world datasets (we redo our analysis excluding it in Appendix~\ref{app:additional_results}). For the pseudo-real datasets, the most commonly used datasets come from the \href{https://www.bnlearn.com/bnrepository/}{bnlearn repository} \citep{scutari2009learning, friedman1997impact}.
    
    These two widely used datasets have some notable limitations. For the Sachs datasets, the consensus network used is not fully consistent with the one given in \citet{sachs2005causal}, in particular, the cycles are often omitted; the ground truth provided is not definitive (see \citet{ramsey2018fask, mooij2020joint}), and varies between studies. Unfortunately, the existence of a ground truth network leads to an over-reliance on structural metrics (less than 5\% of studies use interventional metrics) - even though the dataset includes several interventions. Also, some studies rely on a pseudo-real version of the dataset generated by fitting a model to a consensus network (without cycles) \citep{scutari2009learning}. Finally, the dataset is often not discriminatory for causal discovery methods: the reported performance has peaked at an SHD of around 12. For the pseudo-real datasets from bnlearn, most studies rely only on structural metrics and we note that the way the datasets are generated, all common assumptions are respected.

    Table~\ref{tab:metrics} summarizes the types of metrics used to evaluate performance on simulated (synthetic and pseudo-real) versus real-world datasets. Structural metrics dominate for simulated datasets, where the ground-truth graph is known (100\% of studies use them). In contrast, evaluations on real-world datasets depend more heavily on qualitative assessments (36\% vs. 3\%). Structural metrics are also widely used for real-world datasets (67\%) which can be explained by the overreliance on real-world datasets that contain a ground-truth graph. Both observational and interventional metrics are rare across dataset types, used in fewer than 10\% of studies. Overall, most studies rely solely on structural metrics: 86\% for the simulated data and 54\% for the real-world data. We also note that for real-world datasets that do contain interventions, interventional metrics were not used 89\% of the time.

    In summary, we observed that 1) the choice of datasets could be improved to be closer to realistic settings and 2) most studies rely only on structural or qualitative evaluations. We want to emphasize that the community can readily improve its approach. First, by incorporating a broader range of real-world tasks such as some suggested in the following section. Second, by using interventional metrics as they assess the outcomes we truly care about — namely, the effects of unseen interventions. 

\section{Real-World Datasets: Examples and Unique Challenges} \label{sec:unique_challenges}
    This section highlights three scientific fields—biology, neuroscience, and Earth sciences—that offer numerous real-world datasets for the causal discovery community. We outline key datasets where new methods have been applied and suggest others as promising candidates (see Appendix~\ref{app:datasets} for lists and links to datasets). For each field, we explain the nature of the datasets, their challenges, and some potential opportunities. Finally, we also present some works that expand causal discovery methods by tackling unique challenges of real-world datasets where most common assumptions do not hold.

\subsection{Biology: Biomolecular Networks} \label{sec:biology}
    The field of cellular biology stands out as a key area for applying causal discovery \citep{lagani2016probabilistic, uhler2024building}: first since biologists already analyze cellular processes like metabolism and DNA repair through the lens of networks and pathways \citep{pavlopoulos2011using,alberts2022molecular}, but also because it is driven by recent advances in technology that have enabled the creation of large-scale datasets, often including samples generated under interventional conditions. An enhanced understanding of these networks can shed light on development, disease, and other biological processes \citep{emmert2014gene}. Many technologies now exist for inferring cellular activity, with some focusing on \textit{messenger RNA} (mRNA) levels, others on protein levels, and some targeting entire cell populations (\textit{bulk} methods), while newer methods allow observation of individual cells. One prevalent approach is \textit{single-cell RNA sequencing} (scRNA-seq). Gene-editing techniques like CRISPR \citep{qi2013repurposing} are also of great interest, as they can naturally be framed as interventions. Technologies such as Perturb-seq \citep{dixit2016perturb} combine gene perturbations with single-cell RNA-sequencing, making them ideal for generating datasets well-suited to causal discovery. We discuss in more depth the field-specific characteristics of such data in Appendix~\ref{app:bio}.

    \paragraph{Biomolecular datasets.}
    In Table~\ref{tbl:bio}, we summarize several biological datasets that have been studied in causal discovery, with an emphasis on gene expression datasets.
    Prior to the development of scRNA-seq, the gene expression microarray dataset of \cite{wille2004sparse} was a common testbed for causal discovery and other graphical structure learning algorithms \citep{drton2007multiple,buhlmann2014cam}. Following the development of Perturb-seq, several papers have applied causal discovery to such perturbational gene expression datasets. Often, the datasets have to be preprocessed and only a small subset of the genes are used to perform causal discovery.
    For instance, the complete dataset of bone marrow-derived dendritic cells (BMDCs) from \cite{dixit2016perturb} contains over $\sim$30,000 measurements of 32,777 genes under CRISPR/Cas9 gene deletion perturbations.
    Following the authors' practice, researchers in causal discovery focus on 24 genes that code for highly influential transcription factors and use a shortened version of the datasets that passed a quality control \citep{wang2017permutation, yang2018characterizing,varici2021scalable}.

    More recently, \cite{replogle2022mapping} introduced three much more comprehensive Perturb-seq datasets, including a dataset of 2.5 million K562 cells under thousands of interventions, and two smaller datasets focused on more putatively important genes.
    These datasets have primarily been used for directly predicting the effects of genetic perturbations \citep{lopez2023learning,roohani2024predicting}, but have also been considered in causal discovery \citep{xue2023dotears,lagemann2023deep}.
    Indeed, \cite{chevalley2022causalbench} use these two datasets as the basis of \emph{CausalBench}, a benchmarking suite for causal discovery.
    Lastly, the Perturb-CITE-seq datasets from \cite{frangieh2021multimodal} have been widely used in causal discovery \citep{lopez2022large,sethuraman2023nodags,rohbeck2024bicycle}; these datasets also include protein expression data, which is typically ignored.

    As mentioned in Section~\ref{sec:survey}, the \textit{Sachs} dataset \citep{sachs2005causal} is by far one of the most commonly used real-world datasets in the causal discovery community. \citet{sachs2005causal} pioneered the use of individual cells as the smallest observational unit of intact biological systems, catapulting the available dataset size from dozens or hundreds (of mice, patients, etc) to thousands of cells. 
    This is a flow cytometry dataset that includes abundance measurements for 11 proteins and phospholipids over 7466 CD4+ T cells exposed to nine perturbation conditions. Causal discovery algorithms are often applied to a limited version of the dataset that includes only the 5846 measurements from these seven conditions, see e.g., \cite{wang2017permutation}, \cite{yang2018characterizing}, and \cite{squires2020permutation}.
    The popularity of this dataset is partially accounted for by the fact that, in contrast with the papers above, \cite{sachs2005causal} introduced a consensus network based on existing biological literature. In the absence of such a network, one often resorts to comparing against partial ground truth, e.g., as done by \citet{frot2019robust}, who compare against the reference database TRRUST \citep{han2015trrust}.

    \begin{table}
        \caption{Commonly used biomolecular network datasets. \textit{Int}, \textit{n}, and \textit{d} represent respectively the number of interventions, data points, and features commonly used in causal discovery papers.}
        \small
        \label{tbl:bio}
        \begin{tabular}{ >{\raggedleft\arraybackslash}p{3.83cm}||p{6.8cm}|p{0.6cm}|p{0.9cm}|p{0.6cm}}
         Dataset &Description& Int. & n & d \\
         \hline
         \cite{wille2004sparse} & Gene expression microarray (\textit{A. thaliana}) & - & 118 & 39 \\
         \cite{dixit2016perturb} & Perturb-seq (bone marrow-derived dendritic cells) & 8 & 14427 & 24\\
         \cite{replogle2022mapping} & Perturb-seq (cell line K562) & 1158  & 310385 & 8552\\
        \cite{replogle2022mapping} & Perturb-seq (cell line RPE1) & 651  & 247914 & 8833\\
        \cite{frangieh2021multimodal} & Perturb-CITE-seq (melanoma cells) & 249 & 218331 & 1000\\
        \hline
        \cite{sachs2005causal} & Flow cytometry (CD4+ T cells)  & 6 & 5846 & 11\\
        \end{tabular}
    \end{table}

    \paragraph{Caveats, challenges, and opportunities.}
    We note that while graphs are prevalent in biology, the ``textbook'' examples are significantly different from the kinds of networks learned through causal discovery \citep{tejada2023causal}. 
    In a way, causal sufficiency never holds since biologists typically conceive of networks that involve several different types of molecules, such as membrane channel proteins, enzymes, various other kinds of proteins, and RNA. Meanwhile, causal discovery methods are typically applied to datasets that contain measurements of only a single type of molecule, e.g. gene expression datasets. Thus, the networks returned by typical causal discovery algorithms only explicitly involve genes, though some recent methods are also designed to include other latent factors \citep{squires2022causal,lopez2023learning}. There is also an opportunity to use datasets involving more direct quantification than latent correlates such as mRNA (see Fig~\ref{fig:biology_box}). Technologies such as CRISPR are amazing as they yield many interventional data. We note however that these interventions are often imperfect; in particular, a knocked-out gene may persist in the system, even when it was theoretically completely removed. It is not common practice in the literature to check the perturbations for efficacy, leading to potential issues with both training and validation.

\subsection{Neuroscience} 
\label{sec:neuro}

    Neuroscience is often concerned with understanding mechanisms, which ultimately is about causality \citep{ross2024causation}. It distinguishes the connectome, which describes the wires -- the observable physical connections between neurons or brain areas -- from the effectome, which describes the causal influences between brain regions \citep{pospisil2024fly}. And when it comes to causality, there is a wide spectrum of approaches, including those that assume that correlation is causation and those that ask for perturbations \citep{siddiqi2022causal}. There has been growing interest in how we can uncover genuine causal relationships from neuronal recordings, establishing causal inference as a central paradigm in neuroscience research \citep{reid2019advancing}.

    \textbf{Observational data.} Most causal discovery studies in neuroscience are almost entirely focused on observational data where there is no known ground truth.
    Most branches of neuroscience produce datasets that are used to obtain insights into causal relations. This includes spiking data \citep{stevenson2008inferring}, signals typically recorded at milliseconds resolution of which we currently record about 3000 simultaneously from many brain regions and that is high signal to noise  \citep{stevenson2011advances}. 
    This includes fMRI datasets that are typically recording either about $10^4$ voxels or roughly $10^2$ brain areas at roughly 1Hz resolution \citep{smith2011network}. There are many other modalities including Ca2+ imaging, EEG, MEG, and fNIRS. The key is that there are plenty of datasets available and they are generally either purely observational (and without ground truth causal labels) or come from simulations. We give a list of some frequently used datasets in the field of causal discovery in Appendix~\ref{app:links}. We note that while there are no ground-truth graphs for most datasets, for some, we can rely on the known anatomical connectivity. For example, if there are no anatomical connections, there can not be a direct causal connection \citep{monti2020causal,bird2008hippocampus}.

     \textbf{Challenges and opportunities.} There are major problems for causal inference from brain data. To start with, none of the recording methods obtains data from more than a vanishing subset of underlying variables (e.g., thousands out of many billions of neurons). As such, all observational datasets have dramatically more confounders than observed variables \citep{mehler2018lure}. Many causal inference techniques popular in neuroscience also assume an absence of cycles \citep{friston2011network, zeki1988functional} however, the existence of feedback loops is arguably a key principle of brain connectivity \citep{braitenberg1985charting}. For a more complete list of challenges, see \citet{ramsey2010six, mehler2018lure, stevenson2010similarity, ocker2017statistics, das2020systematic}, who list specific problems of applying causal discovery to brain signals. 
    Recent advancements are also paving the way for performing targeted interventions. Recently, concurrent electrical stimulation with fMRI (es-fMRI) has been proposed \citep{oya2017mapping} and causal discovery, namely fGES, has been applied on such dataset \citep{dubois2020causal}. Combining large-scale perturbations with transcranial magnetic stimulation (TMS) with brain imaging is an interesting avenue to acquire interventional data \citep{oathes2021combining}. Electrical and optogenetic stimulation, which uses light to stimulate genetically modified neurons, is also a promising way to obtain interventional data on animal models \citep{stroh2012optogenetics, lepperod2023inferring, lu2024causal}.  All these studies produce interventional data allowing for a more reliable evaluation of causal discovery methods by verifying if they correctly predict the effects of perturbations. 

\subsection{Earth sciences} \label{sec:earth}
    In the Earth sciences, a field in which controlled experimentation is virtually impossible, researchers rely on a mixture of observational data and physics-based simulations of varying degrees of complexity. Most data is time series or spatio-temporal data and as a consequence, time series causal discovery methods dominate the field, see \citet{runge2019inferring,runge2023causal}. 
    
    \textbf{Reanalysis and observational data.} 
    Due to the intricacies of measuring atmospheric and surface variables across large spatial and temporal scales (e.g., irregular measurement locations or measurement times, meteorological conditions affecting remote sensing capabilities), most studies involving causal discovery in the Earth sciences do not use purely observational data. Instead, the most commonly used type of data, in particular for atmospheric variables, is \emph{reanalysis data}. Reanalysis data is imputed by fitting observations to numerical meteorological prediction models and is thus pseudo-real in the sense of Section~\ref{subsec:pseudoreal}. There are several large reanalysis projects led by national research institutes that make reanalysis data available to the public, including the \href{https://psl.noaa.gov/data/gridded/data.ncep.reanalysis.html}{NCEP/NCAR 40-year reanalysis project} and the \href{https://cds.climate.copernicus.eu}{ERA reanalysis project} \citep{hersbach2020era5}. These databases contain a wide range of atmospheric parameters such as temperature, humidity, pressure, and wind speed direction \citep{kalnay2018ncep}. \citet{runge2019inferring} discuss some of the general challenges of these datasets: strong autocorrelation, time delays, time aggregation, unobserved variables, and more. 
    Examples of causal discovery applications on reanalysis data include \citet{kretschmer2017early, ganesh2023selecting,iglesias2024causally} in which causal discovery is used as a feature selection pre-processing step for downstream prediction tasks and neural network parameter selection. \citet{kretschmer2018different} investigate interactions between global modes of climate variability in the Earth system, so-called teleconnections,  using ERA reanalysis data. \citet{di2020dominant} combine reanalysis data with climate indices available in the \href{https://climexp.knmi.nl/}{KNMI Climate Explorer} to investigate teleconnections in boreal summer; see also \citet{saggioro2020reconstructing} for another causal discovery application to teleconnections using climate indices.
    \citet{di2019long, dicapua2020tropical} use causal discovery on observational data from the \href{https://psl.noaa.gov/data/gridded/tables/precipitation.html}{Climate Prediction Center (CPC) global rainfall dataset} as well as ERA reanalysis surface temperature data to examine causal drivers of Indian summer monsoon rainfall. \citet{engelke2020graphical, amendola2021markov,tran2024estimating} apply causal discovery methods targeting extreme events to a river flow network dataset.
    
    In Environmental Science, causal discovery has been applied in \citet{krich2021functional,krich2022decoupling} to atmospheric flux data from the \href{https://fluxnet.org}{FLUXNET dataset} \citep{pastorello2020fluxnet2015}. FLUXNET contains measurements of carbon, water vapor and energy exchange in different regions of the planet. \citet{guo2024ozone} investigate the influence of ozone levels on influenza with three causal discovery approaches, using data from the \href{https://toar-data.org}{Tropospheric Ozone Assessment Report (TOAR) database}  and the \href{https://www.cdc.gov/flu/weekly/index.htm}{CDC Influenza report}.
    
    \textbf{Physics-based model data.} In addition to direct observations and reanalysis data, climate scientists employ large-scale global or regional climate models to simulate interventions, most notably to investigate global warming under different carbon emission scenarios. Global climate models are coordinated within the \href{https://wcrp-cmip.org}{Climate Model Intercomparison Project (CMIP)}, currently in its 6th phase \citep{eyring2016overview}. However, these simulators are so computationally demanding that it can take months to run a single simulation \citep{balaji2017cpmip}, making it hard to simulate an abundance of interventional data. Additionally, while there is a huge amount of data that has been produced by climate model runs, different datasets are often inconsistent (e.g., due to a different space or time resolution) and may be hard to retrieve. Recently, an effort has been made to make curated versions of these datasets available \citep{watson2022climatebench, kaltenborn2024climateset}. Applications of causal discovery to CMIP data include \citet{karmouche2024changing} who compare the output of the causal discovery method PCMCI+ across different climate models and  \citet{nowack2020causal} who investigate whether CMIP6 models whose causal discovery output graphs are similar to the graph found on reanalysis data exhibit better performance on a downstream prediction task. Simpler data simulators for climate-specific causal discovery that are faster to run but far less detailed have been developed in \citet{ebert2017causal} and \citet{tibau2022spatiotemporal}.

    \textbf{Evaluation of causal discovery output graphs.} Due to the unfeasibility of interventions, it is usually impossible to directly validate the output of a causal discovery method. In addition, as in almost all real-world applications causal discovery assumptions are almost certainly violated, and the degree of violation is often difficult to estimate. Therefore, Earth scientists resort to softer plausibility criteria, for instance by asking whether the returned network is consistent with physical laws. Sometimes more than one causal discovery algorithm is applied to verify whether conclusions are consistent across methods, e.g. in \citet{guo2024ozone}. As Earth scientists are well aware that such validations need to be handled with care due to the danger of confirmation bias, causal discovery is predominantly used for feature selection \citep{kretschmer2017early, ganesh2023selecting,iglesias2024causally} or model comparison \citep{nowack2020causal,karmouche2024changing}.
    
\subsection{Challenges of Real-World Datasets} \label{sec:challenges}
    In this section, we highlight several causal discovery works that have been designed specifically to answer challenges arising in the field of biology and neuroscience. By exploring these works, we aim to illustrate how the violation of standard assumptions can drive innovation, offering insights that purely synthetic or pseudo-real datasets alone might not provide.
    
    \textbf{High-dimensionality.} Real-world datasets often present a high number of features. For instance, brain imaging datasets can contain tens of thousands of features corresponding to individual voxels. \citet{ramsey2017million} proposed fGES, a modification of the popular score-based method GES that assumes linearity, that can scale to a million variables. Gene regulatory network (GRN) datasets, which frequently encompass the entire human genome with around 20,000 features, pose a similar computational challenge, especially for nonlinear causal discovery methods. This is why most applications usually focus on a much smaller subset of genes (often less than a hundred). Recently, a few works have focused on adapting existing nonlinear methods to scale to a much higher number of features (of the order of thousands) \citep{lee2019scaling, lopez2022large}. One specificity of the data that can be leveraged is its modularity. Gene regulatory networks form modules or programs of genes that act together. \citet{segal2005learning, lopez2022large} have used this prior to learning more efficiently causal structures.

    \textbf{Heterogeneity.} The heterogeneity of biological data often necessitates the integration of multiple datasets to achieve a comprehensive understanding of the underlying biological processes. To address this, \citet{triantafillou2015constraint} and \citet{huang2020causal} introduced methods for combining datasets that share a subset of variables, allowing for the leveraging of complementary information across datasets. The heterogeneity can also arise from datasets generated under different populations, such as cell types or disease states. Recognizing this, researchers have proposed methods to model biological data as a mixture of DAGs, each representing a distinct causal structure corresponding to a specific population \citep{saeed2020causal}. Finally, brain imaging datasets are often collected from a cohort of subjects. Although there are strong shared connectivities across the subjects \citep{damoiseaux2006consistent}, each subject also exhibits unique brain connectivity patterns. Exploring methods to conduct multisubject analyses presents a compelling research challenge that has been explored in \citet{oates2014toward, oates2016exact}, \citet{monti2018unified}, and \citet{huang2019specific}.

    \textbf{Cyclic models.} While GRNs and brain connectivity networks contain undoubtedly feedback loops \citep{ferrell2013feedback}, most causal discovery methods assume acyclicity. Recent works motivated by GRNs \citep{rohbeck2024bicycle, sethuraman2023nodags, sethuraman2024learning} and by the brain examples \citep{sanchez2019estimating} have continued the exploration of cyclic causal models.
    
    \textbf{Off-target interventions.} While a gene knockout is usually considered as an intervention targeting a specific gene, in reality, gene knockouts exhibit off-target effects \citep{fu2013high}. In causality terms, this phenomenon is called fat-hand interventions and has been investigated in different biological contexts \citep{eaton2007exact, choo2024causal}.
    
    \textbf{Measurement error.} Technologies such as scRNA-seq can fail to detect some RNA at low levels and will report mistakenly many expression levels at zeros (a phenomenon called dropout) \citep{hicks2018missing}. \citet{saeed2020anchored, ke2023discogen, dai2024gene} have proposed causal discovery methods that take into account this type of measurement error.

\section{Conclusion} \label{sec:conclusion}
    We systematically surveyed recent work in causal discovery research, focusing on datasets and evaluations used in these studies. Our findings reveal that not much has changed since the study of \citet{gentzel2019case}, indicating that the time is well overdue for a critical change in the field. Most studies still only use structural metrics instead of interventional ones. Several studies only include synthetic datasets and while several do include real-world datasets, they often rely on the same ones which have some major limitations. Furthermore, most causal discovery methods rely on strong assumptions that real-world datasets rarely satisfy. Overall, causal discovery still has considerable progress to make before it can be directly applied; practitioners tend to be aware of its limitations and they employ it pragmatically, for instance as an exploratory tool, rather than as a means to derive an irrevocable causal truth. Finally, although we focused on causal discovery, in Appendix~\ref{app:crl} we discuss how similar problems are also present in the emerging field of causal representation learning where simple toy datasets are mostly used and where the common assumptions of the field probably don't hold in real-world settings. We offer recommendations and urge researchers in this field to also use more realistic datasets.

    We also explored in more detail the real-world datasets used in causal discovery. A key observation from our exploration is the increased availability of these kinds of datasets, alongside a trend towards larger and more detailed real-world datasets in recent years. In the field of biology, biomolecular network datasets contain even more interventions than before thanks to new technological advances. These datasets present an invaluable opportunity for the advancement of causal discovery and could also be used in tandem with optimal experimental design as explored in \citet{cho2016reconstructing, ness2017bayesian, agrawal2019abcd, tigas2022interventions, zhang2023active}. Additionally, we showed that real-world domains provide a fertile ground for pushing the boundaries of causal discovery methods since they challenge existing assumptions.
    
    Our conclusion in recommending the use of empirical datasets echoes the one from \citet{gentzel2019case}. To be clear, synthetic datasets are useful, but they should be complemented by more realistic evaluations on pseudo-real and real-world datasets.  When interventional data are present, good quantitative evaluation on real-world datasets exists. However, in many fields besides biology, interventional data are hard to come by and thus pseudo-real datasets might be more adequate. They conserve most of the synthetic datasets' advantages while being more realistic. Still, the creation of pseudo-real datasets should always remain grounded by considering real-world datasets and the assumptions they violate. Through this review, which compiles an extensive list of both simulators and empirical datasets, we aim to motivate researchers to diversify their dataset usage, moving beyond the confines of synthetic data to embrace the complexity and richness of the real world in their causal discovery endeavors.  We hope this effort will also foster the collaboration between causal discovery researchers and practitioners, leading to even more relevant real-world datasets and better integration of domain knowledge.

\appendix

\acks{PB acknowledges the support of the Natural Sciences and Engineering Research Council of Canada (NSERC) and acknowledges Assya Trofimov for helpful discussions. JW received funding from the European Research Council (ERC) Starting Grant CausalEarth (Grant Agreement No. 948112) and from the German Federal Ministry of Education and Research (BMBF) as part of the project MAC-MERLin (Grant Agreement No. 01IW24007). CS received funding from Valence Labs and acknowledges Jason Hartford for helpful discussions. AD acknowledges Sara Magliacane for helpful discussions. KS was funded in part by NIMH grant 1R44MH135465. DS acknowledges support from NSERC Discovery Grant RGPIN-2023-04869, and a Canada-CIFAR AI Chair.}

\bibliography{ref_local}

\newpage
\part{Appendix} %

\section{Systematic Review} \label{app:systematic_review}

\subsection{Methodology}
As explained in the main text we used the Semantic Scholar API to collect papers \citep{semanticscholar}. Specifically, we used the \textit{bulk} method to find scientific articles containing the keywords ``Causal discovery'', ``Causal structure learning'', ``DAG learning'', and ``DAG structure learning'' in their title or abstract. The list of papers was retrieved on September 19, 2024. We did not include articles from workshops at the selected conferences. We manually verified the relevance of each of the 221 papers. We removed articles that were not doing causal discovery or that did not contain any experiments. After this filtering, we kept 167 papers. The whole list of articles and their properties are accessible at our \github \ as a CSV file (\texttt{curated\_papers.csv}).

\subsection{Description of each field}
In this section, we describe briefly each field of the CSV file \texttt{curated\_papers.csv}. The \textit{title}, \textit{year}, and \textit{conf} represent the title of the article, the year it was made accessible (note that this is not necessarily the date of publication if, for example, it was put on an open-access repository such as arXiv), and the name of the conference where the article was published. For the \textit{field} column, we report the field of the provenance of the pseudo-real and real-world datasets. We used the following fields: biology (bio), neuroscience (neuro), Earth Sciences (earth), economy (econ), computational systems (comp), social sciences (socio), health sciences (health), and others. For some common datasets, we noted their use in the \textit{pseudo\_datasets} and \textit{real\_datasets} columns, respectively for pseudo-real and real-world datasets. We noted in \textit{time\_series} and \textit{interv\_setting} if the proposed causal discovery method operated respectively in a time series and/or interventional setting. In \textit{synthetic}, \textit{pseudo\_real}, and \textit{real}, we noted if any experiments were performed on these types of datasets. In \textit{interventions}, we report if the real-world datasets used contained interventional data. In the columns \textit{small}, \textit{medium}, and \textit{big}, we reported the biggest real-world datasets used in each study. Small means 20 variables or less, medium is between 20 and 100 variables, and big is more than 100 variables. The columns \textit{synth\_structural}, \textit{synth\_observational}, \textit{synth\_interventional}, and \textit{synth\_qualitative} correspond to the type of evaluation that was used on synthetic and pseudo-real datasets. By structural, we refer to measures comparing the learned graph to the ground-truth graph such as SHD, SID, AUROC, F1-score, etc. By qualitative, we refer to any qualitative judgment that was done to assess the performance of the algorithm. Most of the time, it was about some edges of the learning graph based on some domain knowledge. By observational and interventional, we refer to metrics such as the negative log-likelihood that evaluate the learned model respectively on held-out data in the observational setting and data in an unseen interventional setting. We give more details of our classification of metrics in Appendix~\ref{app:metrics}. The four following columns (\textit{real\_structural}, \textit{real\_observational}, \textit{real\_ interventional}, and \textit{real\_qualitative}) are similar, but refer to the evaluation on real-world datasets. Finally, \textit{included} denotes whether the article was included or not in our analysis.

\subsection{Scope and limits of the systematic review}
We limited our review to papers published at major machine-learning conferences. Of course, this is not necessarily representative of what practitioners do in their respective fields. The \textit{bulk} method of Semantic Scholar seems adequate for our use as it leads to only a few false positives, but, on the other hand, we might have missed some relevant articles. The choice of categories for the type of datasets was subjectively created on the prevalence of some datasets. Finally, the review was performed by two different reviewers. To alleviate possible bias, the reviewers reviewed a similar subsample to make sure their judgment were similar.

\begin{figure}
  \begin{minipage}[b]{.3\linewidth}
    \centering
    \includegraphics[width=1.\linewidth]{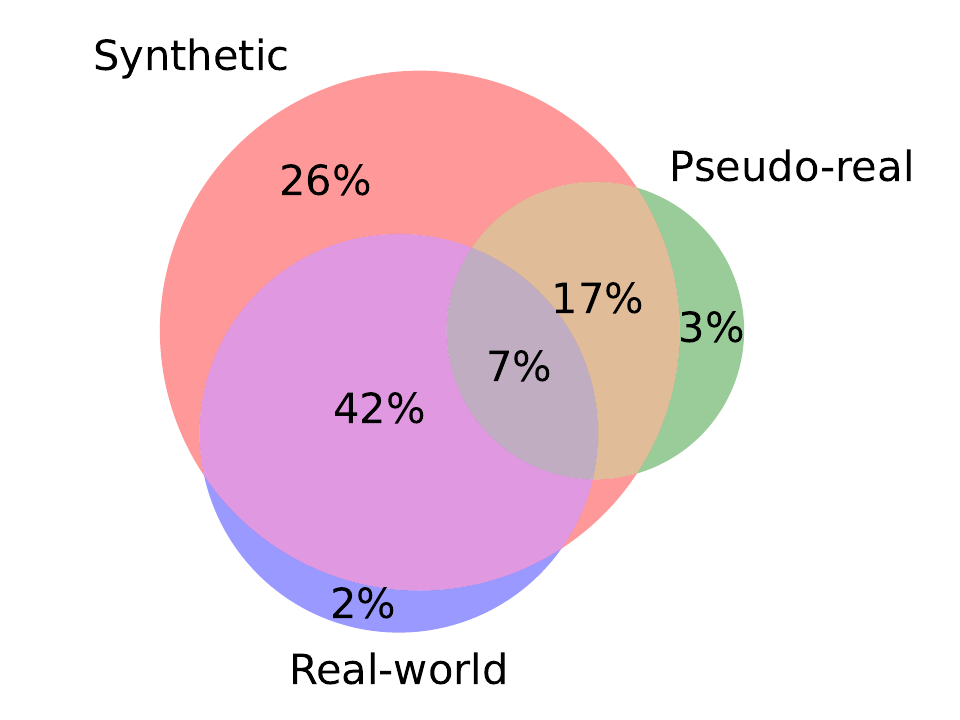} 
    \captionof{figure}{Distribution of papers based on used dataset types.}
    \label{fig:venn_wo_sachs}
  \end{minipage}\hfill
  \begin{minipage}[b]{.3\linewidth}
    \centering
    \includegraphics[width=0.9\linewidth]{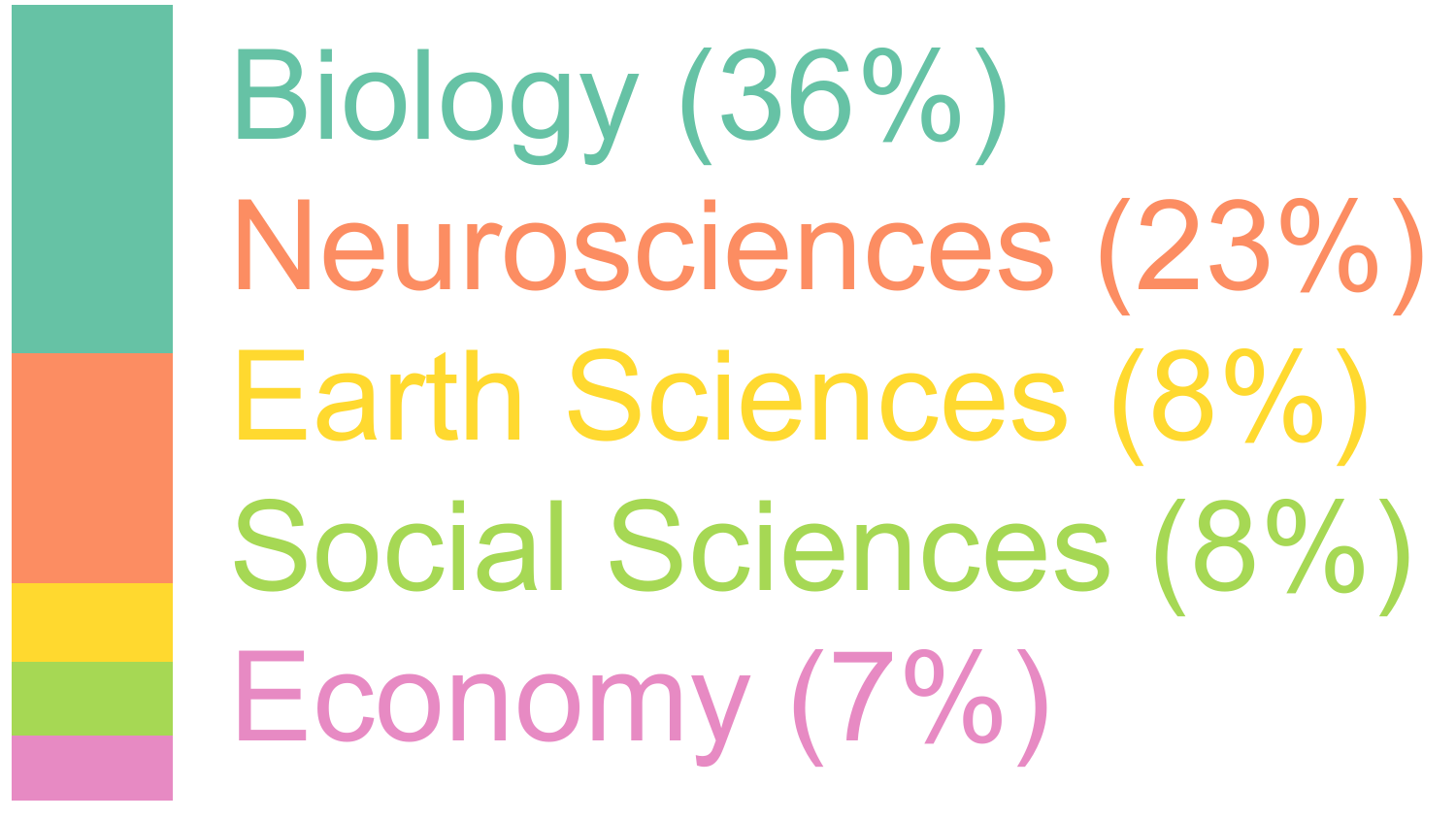}
    \captionof{figure}{Most common fields of the pseudo-real and real-world datasets.}
    \label{fig:fields_wo_sachs}
  \end{minipage}\hfill
  \begin{minipage}[b]{.3\linewidth}
    \centering
    \scriptsize
    \begin{tabular}{l|r|r|r}
        & &Real- \\
        &Simulated &world \\\midrule
        Structural &100.0\% &54.3\% \\
        Qualitative &3.1\% &47.1\% \\
        Observational &5.5\% &7.1\% \\
        Interventional &9.4\% &11.4\% \\
        \bottomrule
    \end{tabular}
    \vspace{5pt}
    \captionof{table}{Percentage of studies using evaluation metrics.} 
    \label{tab:metrics_wo_sachs}
  \end{minipage}\hfill
\end{figure}

\subsection{Additional results} \label{app:additional_results}

\textbf{Excluding papers with only Sachs.} We perform the same analysis as in the main text but we exclude the papers containing only Sachs as their real-world datasets (a total of 37 papers). Overall, the results are similar, but we can notice a few interesting differences. The proportion of papers using only synthetic datasets is higher at 26\% (see Fig.~\ref{fig:venn_wo_sachs}). The field of biology is still the most popular, but it is now more closely followed by the field of neuroscience (see Fig.~\ref{fig:fields_wo_sachs}). Finally, for the real-world evaluations, the use of structural metrics is lower leading to an almost equal use of qualitative and structural metrics (see Table~\ref{tab:metrics_wo_sachs}). This can be explained by the frequent use of Sachs where the structural metrics are used based on the consensus network.

\textbf{Most popular datasets.} In this section, we briefly discuss the most used datasets for real-world and pseudo-real datasets. For the real-world datasets, besides Sachs, the T\"{u}bingen pairs \citep{mooij2016distinguishing} is the most frequent real-world datasets. We describe in more detail this dataset in Appendix~\ref{app:miscellaneous_datasets}. This contains only pairs of variables and the ground truth is assumed to be known, driving up the number of the use structural metrics for real-world datasets. The third most used dataset is the resting state fMRI data from \citet{poldrack2015long}. The recording comes from a single subject over 84 successive days. This is a small graph (6 nodes) representing regions of the hippocampus.  We note that in some studies, the different days are considered as different experimental conditions. The ground-truth graph is unknown and qualitative metrics are mostly used. Finally, in fourth position is the perturb-CITE-seq data from \citet{frangieh2021multimodal} coming from three different cell populations, which contains approximatively 20000 genes (for all studies, only a subset is used $\leq$1000), and interventions under the form of gene knockdowns. The most common metric used is the interventional one.

For the pseudo-real datasets, the bnlearn repository~\citep{scutari2009learning} is followed by the simulated fMRI data from \citet{smith2011network}, the DREAM datasets \citep{marbach2010revealing, greenfield2010dream4}, and SERGIO that generates single-cell expression data of gene regulatory networks from \citet{dibaeinia2020sergio}. For all of them, since they are simulated, the ground-truth graph is known and structural metrics are mostly used. Note that the simulated fMRI data violates the acyclicity assumption by relying on differential equations model.

\section{Biological data} \label{app:bio}
    \subsection{Gene expression and transcriptomics.}
    Although a single organism is composed of vastly different types of cells (e.g., skin cells, neurons, immune cells), all of these cells have the same genetic code (DNA).
    Within an organism, variation in cell state is not driven by variation in genetic profiles.
    Rather, such variation depends heavily upon the process of \textit{transcription}, in which (protein-coding) genes from the DNA are transcribed into messenger RNA (mRNA) molecules, which are then used as a template to synthesize the cell's proteins. One of the most important determining factors of a cell's state is its \textit{transcriptome} (also called its \textit{gene expression profile}), i.e., the total number of mRNA molecules transcribed from each gene.
    Thus, the field of \textit{transcriptomics} is a key part of understanding questions about development, disease, and other processes \citep{emmert2014gene}.

    In causal discovery, one might say, as a shortcut, that \GeneA\ regulates \GeneB\ if changing the expression of \GeneA\ results in a change in the expression of \GeneB. Physically speaking, this relationship is mediated by other, unmeasured molecules, e.g. \GeneA\ might code for a transcription factor (i.e., a protein) which in turn binds to a promoter region for \GeneB, increasing the expression of \GeneB. Thus, in causal discovery, an edge \GeneA $\to$ \GeneB\ represents the existence of such a mediated causal relationship. 
    
    Transcriptomic technologies exist both for measuring and for experimentally manipulating gene expression.
    Two common approaches to measuring gene expression are \textit{microarrays}, which measure a fixed panel of genes, and the much more comprehensive \textit{RNA sequencing (RNA-seq)}, which sequences all mRNA transcripts in an untargeted manner. Molecular measurements are either done in bulk, wherein a population of cells are lysed and an average is measured, or can be performed in single cells. \textit{Single-cell RNA sequencing} (scRNA-seq) has obvious advantages with respect to the number of observational units; however, it should be noted that the data can be extremely sparse. Low abundance genes - including crucial regulators like transcription factors - may fall below the level of detection in individual cells, but are readily detectable in bulk. Single-cell experiments are also far more expensive. 

    \begin{figure}
        \includegraphics[width=0.8\textwidth]{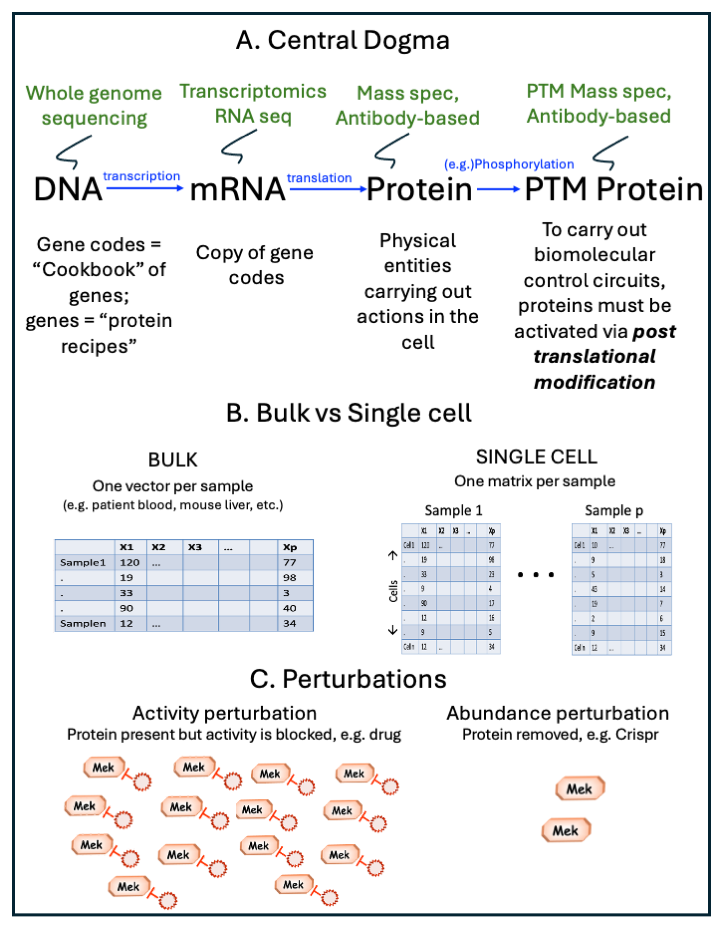}
        \centering
        \caption{Details of biomolecular datasets. A. Central Dogma of biology. The unique code for each organism is encoded in the genome, consisting of a sequence of genes encoded in DNA. The uniqueness is due to variations in genes. Genes are codes or “recipes” for proteins. This code is copied into gene-specific mRNA molecules, via a process called transcription. The information from mRNA molecules is used to create unique proteins via a process called translation. Proteins that comprise the nodes of biomolecular regulatory networks must be activated via processes catalyzed by other (upstream) proteins, in a process called post-translational modification. Processes are shown in blue, measurement technologies are shown in green. Antibody-based modalities include flow cytometry and microscopy-based technologies. B. Depiction of lysate-based (bulk) measurements vs. single cell. Bulk technologies are easier and cheaper, but yield just one vector per sample; single-cell data is sparse in the context of transcriptomics, or of limited dimensionality in proteomics. C. Activity perturbations such as small molecule inhibitors (drugs), and abundance perturbations, typically carried out by genetic means such as CRISPR or ASOs. Activity perturbations leave the protein intact, but block its activity, while abundance perturbations, typically remove the protein by removing the gene from the DNA (genome), or by removing the gene's transcript.}
        \label{fig:biology_box}
        \vspace{-3pt}
    \end{figure}

    \subsection{Beyond transcriptomics.}
    As aforementioned, mRNA readout of gene expression is highly informative of cell state. However, one has to keep in mind that the relation between genes is always mediated: genes themselves do not execute functions in the cells, but rather the proteins which are created based on the information encoded in the genes. Roughly speaking, mRNA (from Gene X) translates into protein (to Protein X). However, factors such as RNA stability and degradation strongly affect this relationship. Hence, the \textit{Gene Regulatory Networks} (GRN) being modeled via transcriptomics are actually carried out not by the quantified gene mRNAs, but by latent variables: the gene-encoded proteins. One way to model biomolecular networks more directly is via proteomic datasets, especially ones in which the abundance of the \emph{activated} proteins is quantified \citep{phosflow,sachs2005causal, PTMmassspec}, though these tend to be more challenging experimentally. 
    
    \textbf{Lysed vs. single cells.} Starting with cells in a dish or test tube, cells are either lysed and measured in bulk, or measured as individual cells. In transcriptomics, single-cell data may be very sparse, as genes may not need to be expressed all the time (if the coded proteins are stable), or may fall below the limit of detection. Bulk data is also far cheaper. In proteomics, mass spectrometry-based modalities measure the entire proteome, but are still in the very early days of single-cell capabilities, and are expensive even in bulk measurements. Most such datasets focus on proteome abundance, some also include \textit{post-translational modification} (PTM). Antibody or label-based modalities for single-cell proteomics such as flow cytometry have been around the longest (>50 years) and are neither sparse nor prohibitively expensive, and readily report an abundance of PTM proteins. However, they must focus on far more limited sets of proteins, in the tens rather than thousands.
    
    \textbf{Activity vs. abundance perturbations.} It is useful to distinguish between activity perturbations such as
small molecule inhibitors (drugs), and abundance perturbations, typically carried out by genetic means such as
CRISPR or ASOs. Activity perturbations leave the protein intact, but block its activity, such that it cannot activate
further proteins in the signaling cascade or biomolecular network, while abundance perturbations typically
remove the protein by removing the gene from the DNA (genome), or by removing the gene’s transcript. Details of data modalities and intervention technologies are summarized in Fig.~\ref{fig:biology_box}.

\section{Datasets} \label{app:datasets}

\subsection{Links to datasets} \label{app:links}
In the following tables, we provide links to the different datasets we discussed in the main text. In Table~\ref{tbl:link_pseudo_datasets}, \ref{tbl:linkbio}, \ref{tbl:link_neuro_datasets} and \ref{tbl:link_earth science} we report the links for pseudo-real, biology, neuroscience and Earth science datasets. 
We also keep an updated version of these lists of datasets at our \github. The lists contain mainly datasets that have been used in causal discovery studies. Several are particularly interesting since they contain interventional data and can violate some common assumptions. Still, we advise thoughtful use in line with our recommendations, rather than applying them blindly. Also, the lists are by no means exhaustive and we do encourage researchers to explore new applications that are not listed here. While this section provides links to pseudo-real and real-world datasets, we would also like to share the synthetic generator \href{https://causally.readthedocs.io/en/latest/}{Causally} \citep{montagna2024assumption} where it is natively possible to control the modelling assumptions (e.g., presence of latent confounders, unfaithful path, etc).

\begin{table} 
\caption{Links to pseudo-real datasets}
\label{tbl:link_pseudo_datasets}
\footnotesize
\centering
\begin{tabular}{ p{4cm}||p{5.7cm}|p{3.8cm}}
     Article & Name & Link \\
     \hline
     \citep{cheng2023causaltime} & CausalTime & \href{https://www.causaltime.cc/}{link to dataset} \\
     \citep{gobler2023causalassembly} & CausalAssembly & \href{https://github.com/boschresearch/causalAssembly}{link to dataset} \\
     \citep{lawrence2021data} & - & \href{https://github.com/causalens/cdml-neurips2020}{link to dataset} \\
    \citep{pratapa2020benchmarking} & BEELINE & \href{https://github.com/murali-group/BEELINE}{link to dataset} \\
    \citep{dibaeinia2020sergio} & SERGIO & \href{https://github.com/PayamDiba/SERGIO}{link to dataset} \\
    \citep{runge2019inferring} & CauseMe & \href{https://causeme.uv.es/}{link to dataset} \\
    \citep{sanchez2019estimating} & - & \href{https://github.com/cabal-cmu/Feedback-Discovery}{link to dataset} \\
    \citep{tu2019neuropathic} & Neuropathic Pain Diagnosis Simulator & \href{https://github.com/TURuibo/Neuropathic-Pain-Diagnosis-Simulator}{link to dataset} \\
    \citep{schaffter2011genenetweaver} & GeneNetWeaver & \href{https://gnw.sourceforge.net/genenetweaver.html}{link to dataset} \\
    \citep{smith2011network} & Netsim & \href{https://www.fmrib.ox.ac.uk/datasets/netsim/}{link to dataset} \\
    \citep{marbach2010revealing} & DREAM4 & \href{https://www.synapse.org/Synapse:syn3049712/wiki/74630}{link to dataset} \\
     
    \citep{van2006syntren} & SynTReN & \href{https://bmcbioinformatics.biomedcentral.com/articles/10.1186/1471-2105-7-43}{link to dataset} \\
\end{tabular}
\end{table}

\begin{table} 
\caption{Links to biology datasets}
\label{tbl:linkbio}
\footnotesize
\centering
\begin{tabular}{ p{3.2cm}||p{7.3cm}|p{3cm}}
     Dataset & Description & Link \\
     \hline
     \citep{replogle2022mapping} & Perturb-seq (cell line K562) & \href{https://github.com/causalbench/causalbench}{link to dataset} \\
    \citep{replogle2022mapping} & Perturb-seq (cell line RPE1) & \href{https://github.com/causalbench/causalbench}{link to dataset} \\
     \citep{frangieh2021multimodal} & Perturb-CITE-seq data from melanoma cells & \href{https://singlecell.broadinstitute.org/single_cell/study/SCP1064/multi-modal-pooled-perturb-cite-seq-screens-in-patient-models-define-novel-mechanisms-of-cancer-immune-evasion}{link to dataset} \\
     \citep{frot2019robust}  & RNA-seq of ovarian cancer & \href{portal.gdc.cancer.gov}{link to dataset} \\
     \citep{dixit2016perturb} & Perturb-Seq of bone marrow-derived dendritic cells & \href{https://www.cell.com/cell/fulltext/S0092-8674(16)31610-5}{link to dataset} \\
     \citep{singer2016distinct} & Naive and activated T cells (Drop-seq) & \href{https://www.cell.com/cell/fulltext/S0092-8674(16)31149-7}{link to dataset} \\
     \citep{sachs2005causal}   & Flow cytometry dataset of immune cells  & \href{https://www.bnlearn.com/research/sachs05/}{link to dataset} \\
     \citep{wille2004sparse} & Microarray of \textit{A. thaliana} gene expression & \href{https://link.springer.com/article/10.1186/gb-2004-5-11-r92#MOESM1}{link to dataset} \\    
\end{tabular}
\end{table}

We provide the links pointing to the original datasets. However, several of these datasets require some data preprocessing. In this paragraph, we elaborate on some datasets giving references to preprocessed versions that have been in causal discovery settings.

The two datasets from \citet{replogle2022mapping} have been processed and curated by \citet{chevalley2022causalbench}, and made accessible on \textit{CausalBench}. The dataset from \citet{frangieh2021multimodal} has been preprocessed by \citet{lopez2022large}. The \href{https://github.com/Genentech/dcdfg/blob/main/perturb-cite-seq/0-data-download.ipynb}{code} is available online. 

For the simulator SERGIO \citep{dibaeinia2020sergio}, a modified \href{https://github.com/haeggee/bacadi/blob/main/sergio/sergio_mod.py}{version} has been proposed by \citet{hagele2023bacadi}, making it possible to use custom graphs and generate (perfect) interventional data.

\begin{table} 
\caption{Links to neural datasets}
\label{tbl:link_neuro_datasets}
\footnotesize
\centering
    \begin{tabular}{p{3.5cm}||p{6cm}|p{1.9cm}|p{1.9cm}}
     Dataset & Description & Link raw data & Preprocessed \\
     \hline
DANDI & Large collection of modern large neuroscience datasets, including optogenetics (DANDI) & \href{https://datadryad.org/stash/dataset/doi:10.5061/dryad.2z34tmpqk}{raw dataset} &\href{https://github.com/dandisets}{link to database} \\
    \citep{dorkenwald2022flywire} & Drosophila connectome (Flywire) & \href{https://codex.flywire.ai}{raw dataset (connectome)} & \href{ https://github.com/dp4846/conn2eff}{simulated Ca2+ activities} \\
     \citep{randi2023neural} & C. elegans simultaneously record Ca2+ most neurons while stimulating & \href{https://osf.io/e2syt/}{raw data} & \href{https://huggingface.co/datasets/qsimeon/celegans_connectome_data}{link to dataset} \\
     \citep{teeters2009crcns} & Spiking data from various sources (CRCNS) & \href{https://crcns.org}{database} & -- \\
     \hline     
     \citep{thompson2020data} & es-fMRI data (intracranial electrodes) & \href{https://openneuro.org/datasets/ds002799/versions/1.0.4}{raw dataset} & same link \\
     \citep{shah2018mapping} & rs-fMRI data from the medial temporal lobe & \href{https://github.com/shahpreya/MTLnet}{raw dataset} & -- \\
     \citep{poldrack2015long} & Hippocampal rs-fMRI (MyConnectome project) & \href{https://openneuro.org/datasets/ds000031/}{raw dataset} & -- \\
     \citep{di2014autism} & rs-fMRI (ABIDE Consortium)  & \href{https://fcon_1000.projects.nitrc.org/indi/abide/}{raw dataset} & \href{http://preprocessed-connectomes-project.org/abide/download.html}{link to dataset} \\
     \citep{van2013wu} & rs-fMRI (Human Connectome Project) & \href{https://db.humanconnectome.org}{raw dataset} & -- \\
     \citep{ramsey2010six} & Task fMRI (Rhyme judgment)  & \href{https://openneuro.org/datasets/ds000003/versions/00001}{raw dataset} & \href{https://github.com/cabal-cmu/Feedback-Discovery}{link to dataset} \\
     \citep{wang2003training} & Task fMRI (star/plus experiment) & -- & -- \\
    \end{tabular}
\end{table}

\begin{table} 
\caption{Links to Earth science datasets}
\label{tbl:link_earth science}
\footnotesize
\centering
\begin{tabular}{ p{3.5cm}||p{6.9cm}|p{3cm}}
     Article & Name & Link \\
     \hline
     \citep{kaltenborn2024climateset} & ClimateSet & \href{https://climateset.github.io/}{link to dataset} \\
     \citep{nguyen2024climatelearn} & ClimateLearn & {contained in ClimateSet} \\
      \citep{subramaniamclimsim} & ClimSim & {contained in ClimateSet} \\
      \citep{watson2022climatebench} & ClimateBench & {contained in ClimateSet} \\
      \citep{rasp2020weatherbench} & WeatherBench & {contained in ClimateSet} \\
      \citep{de2021rainbench} & RainBench & {contained in ClimateSet} \\
      -- & KNMI Climate Explorer & \href{https://climexp.knmi.nl/}{link to database} \\
     -- & NCEP/NCAR 40-year reanalysis project & \href{https://psl.noaa.gov/data/gridded/data.ncep.reanalysis.html}{link to database} \\
     \citep{hersbach2020era5} & ERA reanalysis project &\href{https://cds.climate.copernicus.eu}{link to database} \\
     -- & Climate Prediction Center (CPC) global rainfall dataset &\href{https://psl.noaa.gov/data/gridded/tables/precipitation.html}{link to dataset} \\
     \citep{pastorello2020fluxnet2015} & FLUXNET & \href{https://fluxnet.org}{link to database} \\
     -- & Tropospheric Ozone Assessment Report (TOAR) & \href{https://toar-data.org}{link to database} \\
     -- & CDC Influenza report & \href{https://www.cdc.gov/fluview/overview/fluview-interactive.html?CDC_AAref_Val=https://www.cdc.gov/flu/weekly/fluviewinteractive.htm}{link to dataset} \\
       \citep{eyring2016overview} & Climate Model Intercomparison Project (CMIP) & \href{https://wcrp-cmip.org}{link to database}\\
      
\end{tabular}
\end{table}

\begin{table} 
\caption{Links to Miscellaneous Real-World Datasets}
\label{tbl:link_miscellaneous}
\footnotesize
\centering
\begin{tabular}{ p{3.5cm}||p{6.9cm}|p{3cm}}
     Article & Name & Link \\
     \hline
     \citep{gamella2024causal} & Causal Chambers & \href{https://github.com/juangamella/causal-chamber}{link to simulator} \\
     \citep{wang2022learning} & MOS 6502 microprocessor & \href{https://github.com/KordingLab/LearningCausalDiscovery}{link to dataset} \\
     \citep{gentzel2019case} & Software Systems (PostgreSQL, JDK, Networking) & \href{https://groups.cs.umass.edu/kdl/causal-eval-data/}{link to datasets} \\
     \citep{huang2018generalized} & Archaeology data set & -- \\
     \citep{mooij2016distinguishing} & Cause-effect pairs (Tübingen pairs) & \href{https://webdav.tuebingen.mpg.de/cause-effect/}{link to dataset} \\
     \citep{byrne2013structural} & Teacher’s burnout study & \href{https://github.com/kurowasan/cd_datasets/blob/main/data/teacher_burnout_data.csv}{link to dataset} \\
     \citep{shimizu2011directlingam} & General Social Survey & \href{https://github.com/kurowasan/cd_datasets/blob/main/data/general_status_survey.csv}{link to dataset} \\
      
\end{tabular}
\end{table}

\subsection{Miscellaneous Datasets} \label{app:miscellaneous_datasets}
    In this section, we elaborate on a few other source of datasets that are particularly interesting.
    
    \textbf{Computational systems.} Datasets generated by existing computational systems have been recently proposed and used to evaluate causal discovery methods. They are of considerable interest since they are deterministic yet complex systems composed of many components. Also, compared to pseudo-real datasets, they are generated from real-world environments. For example, the C++ simulator of the MOS 6502 microprocessor  \citep{jonas2017could, wang2022learning} is composed of 3510 transistors and 1904 connection elements where the variables of interest are the voltage of the different transistors. While the physical connections are known, it is not sufficient to know the causal graph, instead, it was determined from the perturbation of single transistors. Similarly, data from the analysis of microservice-based applications \citep{ikram2022root} (from a sock-shop demo and a production-based microservice system hosted on AWS cloud-native system) has been used to evaluate root-cause analysis methods. The data can be different metrics such as CPU and memory utilization, while the interventions can be failures such as CPU hog and memory leak. \citet{gentzel2019case} also proposed several datasets from such systems: Java Development Kit, PostgreSQL, and a web server infrastructure. Recently, datasets generated by the activation of neural networks have also been used in order to interpret their learned representations \citep{geiger2021causal, geiger2022inducing}.
    
    \textbf{The Causal Chambers.} Recently, \citet{gamella2024causal} have proposed an interesting new type of dataset where the data is generated from real-world experiments but the ground-truth graph is known. \citet{gamella2024causal} designed two computer-controlled physical simulators, that can generate observational and interventional data (i.i.d. as well as time series) with experimentally verified ground-truth graphs. Due to the recency of its development, as far as we know, no causal discovery method has been evaluated on Causal Chamber data yet (see \citet{lohse2024sortability} for an investigation on var-/$R^2$-sortability of Causal Chambers time series data).

    \textbf{T\"{u}bingen pairs.} The \href{https://webdav.tuebingen.mpg.de/cause-effect/}{T\"{u}bingen pairs} (or CauseEffectPairs benchmark) \citep{mooij2016distinguishing} is a repository regrouping 108 datasets composed of pairs of cause and effect coming from many different domains. The range of domains is vast: climate data, biology (e.g., growth of abalone), healthcare (e.g., arrhythmia and diabetes), economy (e.g., census income), stock market,  etc. Note that many of these pairs are adapted from the UCI Machine Learning Repository \citep{bachuci} where the complete datasets are available. While being real-world data, the ground-truth causal direction is given by the authors since in many cases, it is obvious from common sense (e.g., the altitude causes the temperature). %
    
    \textbf{TableShift.} Recently \citet{gardner2024benchmarking} proposed \href{https://github.com/mlfoundations/tableshift}{TableShift}: a benchmark comprising many tabular datasets chosen to assess the robustness of machine learning methods to distribution shift. Although not originally intended for causal discovery, it has been explored in \citet{nastl2024causal}, where causal discovery methods are used with the aim of finding causal parents of variables of interest.

\section{Metrics to evaluate causal models} \label{app:metrics}
    \textbf{Structural metrics.} The most commonly used structural metric is clearly the structural Hamming distance (SHD) \citep{acid2003searching, tsamardinos2006max}. The distance $\text{SHD}(\hat{G}, G)$ is defined as the number of edges that should be added, removed or reversed in order to modify an estimated graph $\hat{G}$ to a target graph $G$.
    Besides SHD, other similar metrics are also often reported: precision-recall, false discovery rate, F$_{1}$ score, AUROC, etc. They can be more useful since the SHD alone can be misleading (e.g., for a really sparse graph, an empty graph can be better in terms of SHD than denser graphs that contain many ground-truth edges). 
    
    A major limitation of these metrics is that they are purely based on the graph without any notion of causality.
    Other structural metrics assess the distance in terms of topological ordering (e.g., \citet{rolland2022score}) conditional independencies \citep{textor2016robust}, d-separation statements \citep{wahl2024metrics}, node-permutation tests \citep{eulig2023toward}, etc. More focused on the effect of interventions, \citet{peters2015structural} proposed the \textit{Structural Interventional Distance} (SID) which counts the number of interventional distributions that would be wrongly computed using the parents from the learned structure as its adjustment sets. While considering interventions, this measure is still about the graph and correlates strongly with SHD \citep{gentzel2019case}. A generalization of this measure that considers other adjustment sets has also been proposed by \citet{henckel2024adjustment}. It also has the advantage of being directly applicable to CPDAGs as it returns a scalar instead of bounds and is computationally less demanding.

    \textbf{Interventional metrics.} As previously mentioned, interventional metrics do not necessitate a known ground-truth graph as it evaluates directly how well a causal model can predict data coming from an unseen interventional distribution \citep{garant2016evaluating}. A common interventional metric is the \textit{interventional negative log-likelihood} (I-NLL) \citep{lopez2022large}:
        \begin{equation}
            \text{I-NLL} = -\mathbb{E}_{x \sim P^{(j)}}[\log P_\theta^{(j)}(X)],
        \end{equation}
    where data from $P^{(j)}$ were not part of the training set and $P^{(j)}_\theta$ is the learned model. Usually, the average is taken over multiple interventional distributions. We also note that this metric does not even require the learned model $P_\theta$ to use a graph and can thus be used with a more general class of methods.  While it often takes the form of a likelihood, it can also be any distance between the learned distribution and the ground-truth one: some have used the mean absolute error \citep{lopez2022large}, total variation distance \citep{garant2016evaluating}, the KL divergence, etc. It can also take the form of the strength of a causal relation or the average/conditional treatment effect.

\section{Causal Representation Learning} \label{app:crl}
    In this review, we focused on causal discovery where it is assumed that we have access to structured data. However, many datasets generated by real-world phenomena are unstructured data (e.g., images, videos, texts, etc). The question of how to deal with such datasets has been central in causal abstraction \citep{rubenstein2017causal, beckers2019abstracting, beckers2020approximate, massidda2023causal}, causal feature learning \citep{chalupka2014visual, chalupka2017causal}, causal grouping \citep{parviainen2017learning,anand2023causal,wahl2023vector,wahl2024foundations}, and causal representation learning \citep{scholkopf2021toward}. In this section, we will briefly present causal representation learning where the main task is identifying latent causal variables usually from an unstructured input. This recent development opens the doors to many new practical applications where datasets are unstructured. So far, the field has focused on proving identifiability results, showing that it is possible to recover the right representation (up to some minor transformations) under some assumptions.
    
    Formally, we have the observed variable $X = (X_1, \dots, X_n)$ that are generated by applying a function $g$ to the latent variable $Z = (Z_1, \dots, Z_d)$ that is Markov to a graph $G$. The data-generating process is as follows:
    \begin{equation} \label{eq:crl}
        X = g(Z), \ \ \ \ \ \ P_Z = \prod_{i = 1}^d P_i(Z_i \mid pa_i^G),
    \end{equation}
    where $g: \mathbb{R}^d \rightarrow \mathbb{R}^n$ is an injective function called decoder or mixing function. The goal is to find from $X$ a representation of $Z$ that is causally disentangled. Disentangled representations allow interpretability and can also be useful for many downstream tasks. However, without assumptions, it is impossible to learn such a representation \citep{hyvarinen1999nonlinear, locatello2019challenging}. Thus, data-generating processes considered follow additional assumptions: assumptions are made on the distribution of the latent variable $Z$ and its support, additional assumptions, such as sparsity, are made about $g$ \citep{moran2021identifiable, rhodes2021local, zheng2022on, brouillard2024causal} or the latent graph~\citep{lachapelle22a} and often the presence of auxiliary information is assumed (e.g., \citet{locatello2020weakly, vonkugelgen2021selfsupervised, brehmer2022weakly, ahuja2022weakly, lachapelle2022synergies}). Given these assumptions, many identification results have been discovered showing that the disentangled representation is unique up to some minor transformation (such as affine transformations). 
    
\begin{figure}[t] 
        \includegraphics[width=0.95\textwidth]{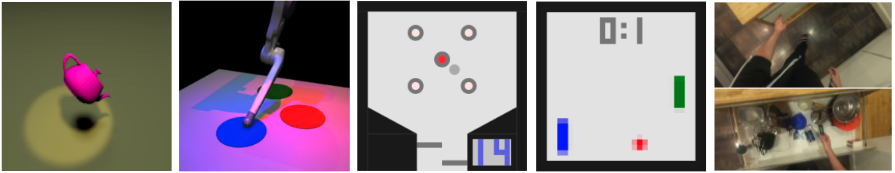}
        \centering
        \caption{From left to right: Causal3DIdent, CausalCircuit, Causal Pinball, Interventional Pong, Causal triplet (see Table~\ref{tbl:crl} for the references).}
        \label{fig:crl_datasets}
    \end{figure}

    However, so far, the field has compared new methods almost exclusively on simple synthetic datasets. We present in Table~\ref{tbl:crl} a list of commonly used datasets in causal representation learning and show visual examples in Figure~\ref{fig:crl_datasets}. These synthetic datasets are, in many respects, really not representative of real-world tasks: they focus only on problems where images are the observable input, the latent variables are always simple properties of objects (e.g., position, color, etc), and the latent variables are only a few (i.e., less than 10).  \citet{liu2023causal} also highlight that images coming from synthetic datasets are too simple: most have plain textures, contain only a small number of objects, and do not contain object occlusion (see Figure~\ref{fig:crl_datasets}, except the image to the right). As for the evaluation metrics, the situation is similar to causal discovery. Most rely on the Mean Correlation Coefficient (MCC) that finds the best permutation to evaluate how well the learned latent variables correlate with the ground-truth latent variables. However, the MCC is, at least in some instances, not directly related to the models' performance on downstream tasks. As for the causal discovery evaluation, we recommend evaluating models on downstream tasks such as predicting the effect of interventions.
    
    \begin{table}
         \caption{List of the most common CRL datasets. $d_z$ is the dimensionality of the latent variables.}
         \label{tbl:crl}
         \small
         \begin{tabular}{ p{4cm}||p{7.7cm}|p{0.6cm}p{1cm}}
         Dataset &Description & $d_z$ \\
         \hline
         \citet{von2021self}   & Causal3DIdent: a 3D object under various conditions & $7$\\
         \citet{brehmer2022weakly} & CausalCircuit: A robot arm  can interact with lights & 4\\
         \citet{lippe2022causal} & Causal Pinball & 5\\
         \citet{lippe2022citris} & Interventional Pong & 6\\
         \citet{liu2023causal} & Causal triplet  & -\\
        \end{tabular}
    \end{table}

    A few recent works have used real-world datasets, such as \citet{lopez2023learning, zhang2024identifiability} for gene regulatory networks, \citet{yao2024marrying, brouillard2024causal} in the Earth science domain, and \citet{cadei2024smoke} in ecology. However, no realistic simulators like the one proposed for causal discovery exist.  We also observe that many real-world datasets reported in Table~\ref{tbl:linkbio}, \ref{tbl:link_neuro_datasets} and \ref{tbl:link_earth science} are good candidates for causal representation learning methods since they are high-dimensional unstructured data before their preprocessing. These datasets offer a more diversified and challenging repertoire than what is presently used in the field. Furthermore, the common practice in causal discovery applied to unstructured problems is to use dimensionality reduction methods or to drop features with less variation. This constitutes an opportunity for causal representation learning since these common practices probably lead to an incorrect choice of variables. 
    
    We conclude by stating that we only focused on causal representation learning, but the realm of domains where causal methods are applied has grown abundantly yielding many other possible applications. While deviating from classical causal discovery, they often use more realistic datasets. Some new methods relying on LLM can directly rely on text metadata to learn relevant causal variables \citep{liu2024discovery}. Causally-inspired algorithms have also been proposed to tackle the problem of multidomain data (e.g., \citet{arjovsky2019invariant}), where different domains are interpreted as different interventional environments (for examples of datasets, see \citet{gulrajani2020search}). Finally, we could also mention the active field of causal reinforcement learning (e.g., \citet{ahmed2020causalworld, ke2021systematic}).
\end{document}